\documentclass[conference]{IEEEtran}

\usepackage{xspace}

\def\0{{\mathbf  0}}
\def\1{{\mathbf  1}}

\newlength\savewidth

\newcommand{\discriminator}{Side-view Normal Discriminator\xspace}

\newcommand{\ld}{\textcolor{black}}

\newcommand{\sexyname}{CHRIS\xspace}
\usepackage{booktabs}
\def\eg{\emph{e.g.,}} 
\def\ie{\emph{i.e.,}}

\def\wrt{{w.r.t.}} 
\def\etal{\emph{et al.~}}
\newcommand{\ourng}{Multi-to-one Gradient Computation\xspace}
\newcommand{\ourdis}{Side-View Normal Discriminator\xspace}
\newcommand{\ourvoxel}{\mathcal{M}_v^{3D}}

\usepackage{pifont}
\usepackage[dvipsnames]{xcolor}
\usepackage{lipsum}
\usepackage{color, colortbl}
\usepackage{amsmath}
\usepackage{amsthm}
\usepackage{array}
\usepackage{newlfont}
\usepackage{textcomp}
\usepackage{multirow}
\usepackage{commath}
\usepackage{hhline}
\usepackage{tabularx}
\usepackage{bigstrut}
\usepackage[para,online,flushleft]{threeparttable}
\usepackage{etoc}
\usepackage[ruled,linesnumbered]{algorithm}
\usepackage{amsthm}

\IEEEoverridecommandlockouts
\usepackage{cite}
\usepackage{amsmath,amssymb,amsfonts}
\usepackage{algorithmic}
\usepackage{graphicx}
\usepackage{textcomp}
\usepackage{xcolor}
\def\BibTeX{{\rm B\kern-.05em{\sc i\kern-.025em b}\kern-.08em
    T\kern-.1667em\lower.7ex\hbox{E}\kern-.125emX}}
\begin{document}

\title{CHRIS: Clothed Human Reconstruction with Side View Consistency\\
\thanks{*Equal Contribution. $^\dag$ Corresponding Author.}
}

\author{\IEEEauthorblockN{1\textsuperscript{st} Dong Liu*}
\IEEEauthorblockA{\textit{South China University of Technology} \\
Guangzhou, China \\
sesmildong@mail.scut.edu.cn}\\

\IEEEauthorblockN{4\textsuperscript{th} Yuxin Gao}
\IEEEauthorblockA{\textit{South China University of Technology} \\
Guangzhou, China \\
202311093534@mail.scut.edu.cn}

\and
\IEEEauthorblockN{2\textsuperscript{nd} Yifan Yang*}
\IEEEauthorblockA{\textit{South China University of Technology} \\
Guangzhou, China \\
seyoungyif@mail.scut.edu.cn}
\and
\IEEEauthorblockN{3\textsuperscript{rd} Zixiong Huang}
\IEEEauthorblockA{\textit{South China University of Technology} \\
Guangzhou, China \\
sesmilhzx@mail.scut.edu.cn}\\

\IEEEauthorblockN{5\textsuperscript{th} Mingkui Tan$^\dag$}
\IEEEauthorblockA{\textit{South China University of Technology} \\
Guangzhou, China \\
mingkuitan@scut.edu.cn}

}

\maketitle

\begin{abstract}
Creating a realistic clothed human from a single-view RGB image is crucial for applications like mixed reality and filmmaking. Despite some progress in recent years, mainstream methods often fail to fully utilize side-view information, as the input single-view image contains front-view information only. This leads to globally unrealistic topology and local surface inconsistency in side views. To address these, we introduce Clothed Human Reconstruction with Side View Consistency, namely CHRIS, which consists of 1) A Side-View Normal Discriminator that enhances global visual reasonability by distinguishing the generated side-view normals from the ground truth ones; 2) A Multi-to-One Gradient Computation (M2O) that ensures local surface consistency. M2O calculates the gradient of a sampling point by integrating the gradients of the nearby points, effectively acting as a smooth operation. Experimental results demonstrate that CHRIS achieves state-of-the-art performance on public benchmarks and outperforms the prior work.


\end{abstract}


\begin{figure}[!th]
    \centering
\includegraphics[width=0.81\linewidth]{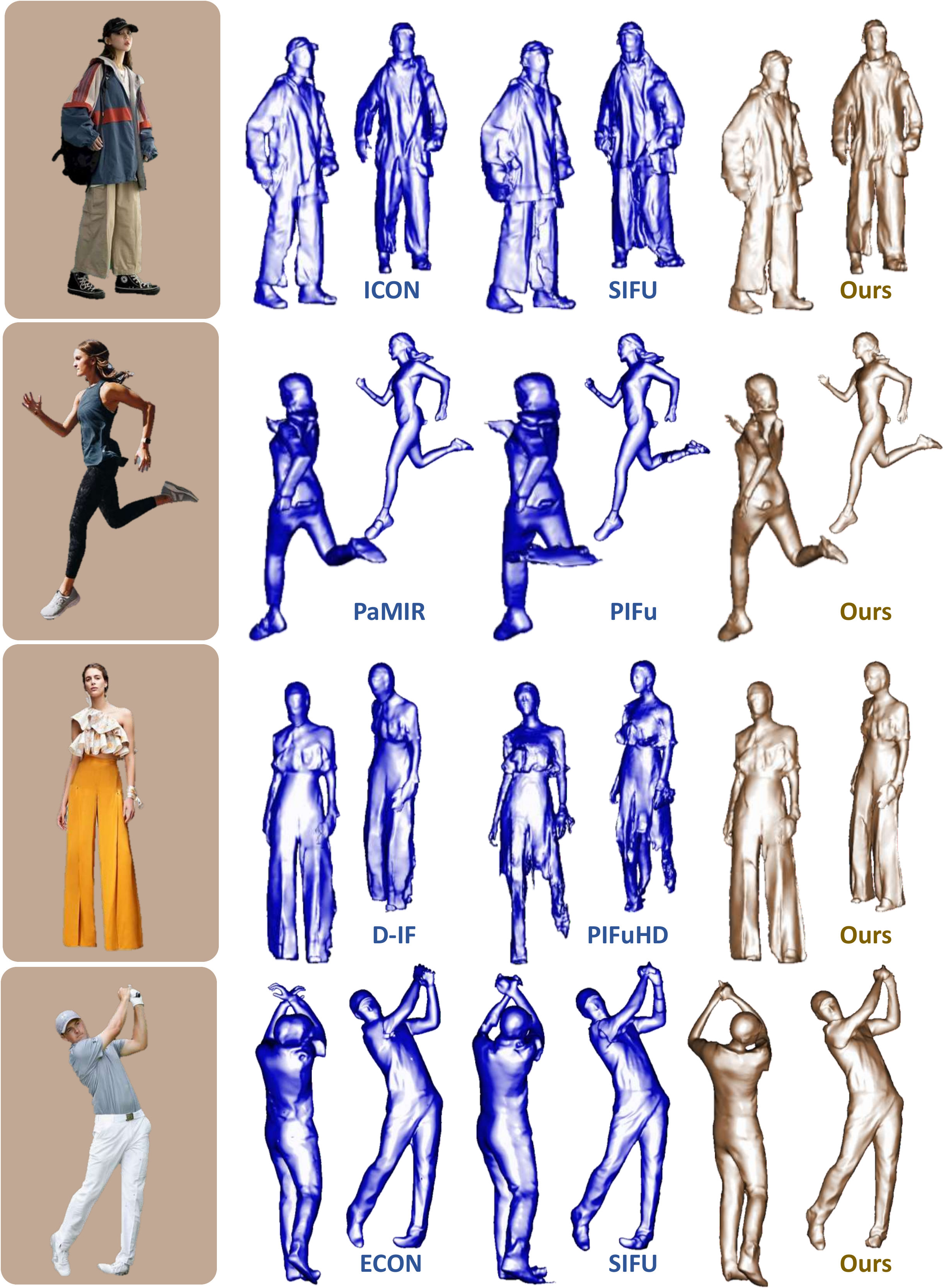}
    \caption{Comparisons with recent advances in in-the-wild images, our \sexyname~achieves more accurate and visually reasonable reconstruction of challenging poses and diverse clothes. Zoom in for more details.}
    \label{fig:vis_comp}
\end{figure}

\section{Introduction}
\label{sec:intro}

Creating realistic digital humans with intricate clothing details plays a pivotal role in the field of mixed reality~\cite{yoon2019effect}
and filmmaking~\cite{sun2022research}. Conventional methods usually require substantial human resources and expensive specialized equipment to customize high-fidelity 3D digital avatars~\cite{schonberger2016pixelwise, guo2019relightables, zheng2014patchmatch} from multi-view images. These manners are cost-ineffective and labor-intensive. 

To simplify the reconstruction process while maintaining quality, recent approaches~\cite{alldieck2019tex2shape, saito2019pifu, saito2020pifuhd, zheng2021pamir, xiu2022icon, xiu2023econ, yang2023dif} extract 3D data from RGB images of clothed human, eliminating the need for costly scanning equipment and making it easier for individuals to create personalized avatars. According to whether involving an implicit function to represent a 3D human model, existing methods can be divided into the following two main categories. 

\textit{Explicit-Shape-Based} methods~\cite{alldieck2019tex2shape, xiang2020monoclothcap} use parametric human body models like SMPL~\cite{loper2015smpl} and SMPL-X~\cite{pavlakos2019smplx} to represent shape and pose of a naked body and then incorporate clothing via 3D offsets or adjustable garment templates. However, they struggle with complex surfaces like skirts and dresses due to the limitations of parametric models.

\textit{Implicit-Function-Based} methods use a deep implicit function only~\cite{saito2019pifu,saito2020pifuhd} or combine the parametric models with implicit functions~\cite{xiu2022icon,yang2024hilo,zhang2024sifu} to enhance reconstruction. A common manner of these methods is that they sample points in 3D space, and feed the 2D/3D features related to the points to the implicit function to infer geometry. Although implicit-function-based methods are effective in handling clothed humans in various scenarios, they do not fully address the problem of global visually reasonable and local surface-consistent reconstruction in the \textbf{side view}. 

 Specifically, PIFu~\cite{saito2019pifu} aligns 2D pixels of the given front-view image with the global 3D human context.    ICON~\cite{xiu2023econ} uses the front view image to infer SMPL-X, obtaining local 3D features from SMPL-X, and uses a deep implicit function to infer geometry.   HiLo~\cite{yang2024hilo} extract high-frequency and low-frequency information from SMPL-X to recover geometry details and improve reconstruction robustness. Recently, SIFU~\cite{zhang2024sifu} use SMPL-X to render side-view normal maps, but these maps represent the naked body rather than the clothed human.  These methods produce high-quality geometry in the front and back views. However, due to the underutilization of side view information of clothed humans, from Fig.~\ref{fig:vis_comp}, they usually achieve global artifacts such as wrong thickness, and local artifacts like stitches, and broken topology in the side views. 

Given the limited supervision of side-view information, it is essential to regularize the side views of the clothed human on both \textit{global} and \textit{local scales} to ensure a visually reasonable and surface-consistent outcome. The global scale regularization focuses on making the full-body side-view geometry visually reasonable, even \textit{without corresponding ground truth side-view information of clothed humans}. Moreover, to address the local inconsistency that leads to unnatural artifacts, we seek to impose a local scale regularization to achieve surface consistency on the local scale.

To achieve this, we propose a human reconstruction paradigm that promotes both global and local side-view consistency. At the global scale, we first render side-view normal maps from the reconstructed mesh. To enhance the quality of side-view geometry, we introduce a~\ourdis that distinguishes the rendered maps from ground truth normal maps. Additionally, to improve local surface consistency, we design a multi-to-one gradient computation along with a coarse-to-fine strategy. By integrating the gradients of nearby points around a sampling point, we smooth local irregularities and ensure a consistent surface.

We qualitatively and quantitatively evaluate our method on in-the-wild images from the internet, THuman2.0, and CAPE datasets. Experimental results verify our superiority over previous
approaches, especially in: 1) global visual plausibility in the side view (see Fig.~\ref{fig:ablation_dis}). 2) the local surface consistency (see Fig.~\ref{fig:ablation_m2o}). Our \sexyname outperforms the state-of-the-art methods in terms of Chamfer and P2S metrics. 3) Our reconstructed humans are animatable.

\textbf{Our contributions}:
    \textbf{1)} We find that the underutilization of side-view regularization in existing methods often leads to global and local artifacts, especially in side views. To address this, we introduce a side-view consistent paradigm that regularizes side-view geometry at both global and local scales.
    \textbf{2)} To regularize global side-view geometry, we introduce a~\ourdis that distinguishes rendered side-view normal maps from ground truth normal maps. Experimental results show that our approach reconstructs a visually reasonable clothed human.
    \textbf{3)} To regularize local side-view geometry, we introduce a multi-to-one gradient computation strategy. By integrating gradients from nearby points, this approach smooths out local irregularities, resulting in a consistent surface. Empirically, we find that this method alleviates local artifacts such as stitches and jagged edges.

\section{Preliminaries}
\label{SEC:PREL}

\noindent\textbf{SMPL and SMPL-X.} The Skinned Multi-Person Linear model (SMPL)~\cite{loper2015smpl} is a skinned vertex-based model that accurately represents a wide variety of body shapes with different human poses.
Its body mesh $\mathcal{M}$ is defined as follows:
\begin{equation} \label{eqn:smpl}
\begin{aligned}
\mathcal{M}(\mathbf{\beta}, \mathbf{\theta}):\mathbb{R}^{|\mathbf{\theta}|\times|\mathbf{\beta}|}~\mapsto~\mathbb{R}^{3N}
\end{aligned}
\end{equation}
Here, $\mathbf{\beta}$ is a vector of shape parameters, $\mathbf{\theta}$ is a vector of pose parameters, and 
$N=6890$ denotes the number of vertices. 
SMPL-X~\cite{pavlakos2019smplx} is an extended version of SMPL that includes fully articulated hands and an expressive face.

\noindent \textbf{Implicit Function} is widely used in modeling complex topology.
We employ an implicit function $\mathcal{\phi}$ to estimate a Signed Distance Function (SDF)~\cite{park2019deepsdf}, which represents the distance of the reconstructed clothed human surface.  Following \cite{yang2024hilo}, given a set of sampling points $\mathbf{P}$, the pipeline of our \sexyname follows the equation to obtain the reconstructed SDF $\hat{\mathcal{S}}(\mathbf{p})$ :
\begin{equation}
\mathrm{\sexyname} : \mathcal{\phi}(\mathcal{F}_{sdf}(\mathbf{p}), \mathcal{F}_{\mathrm{n}}(\mathbf{p}), \ourvoxel({\mathbf{p}})) \rightarrow \hat{\mathcal{S}}(\mathbf{p})
\end{equation}
where $\mathcal{F}_{sdf}(\mathbf{p})$ is the signed distance, $\mathcal{F}_{\mathrm{n}}(\mathbf{p})$ is the concatenation of estimated SMPL surface normal $\mathcal{F}_{\mathrm{n}}^{\mathrm{b}}(\mathbf{p})$ and the normal vector $\mathcal{F}_{\mathrm{n}}^{\mathrm{c}}(\mathbf{p})$, and $\ourvoxel({\mathbf{p}})$ is the voxelized feature.

\section{Clothed Human Reconstruction with Side-view Consistency} 
\label{sec:pghfsdf}

\begin{figure*}[t]
    \centering
    \includegraphics[width=0.9\linewidth]{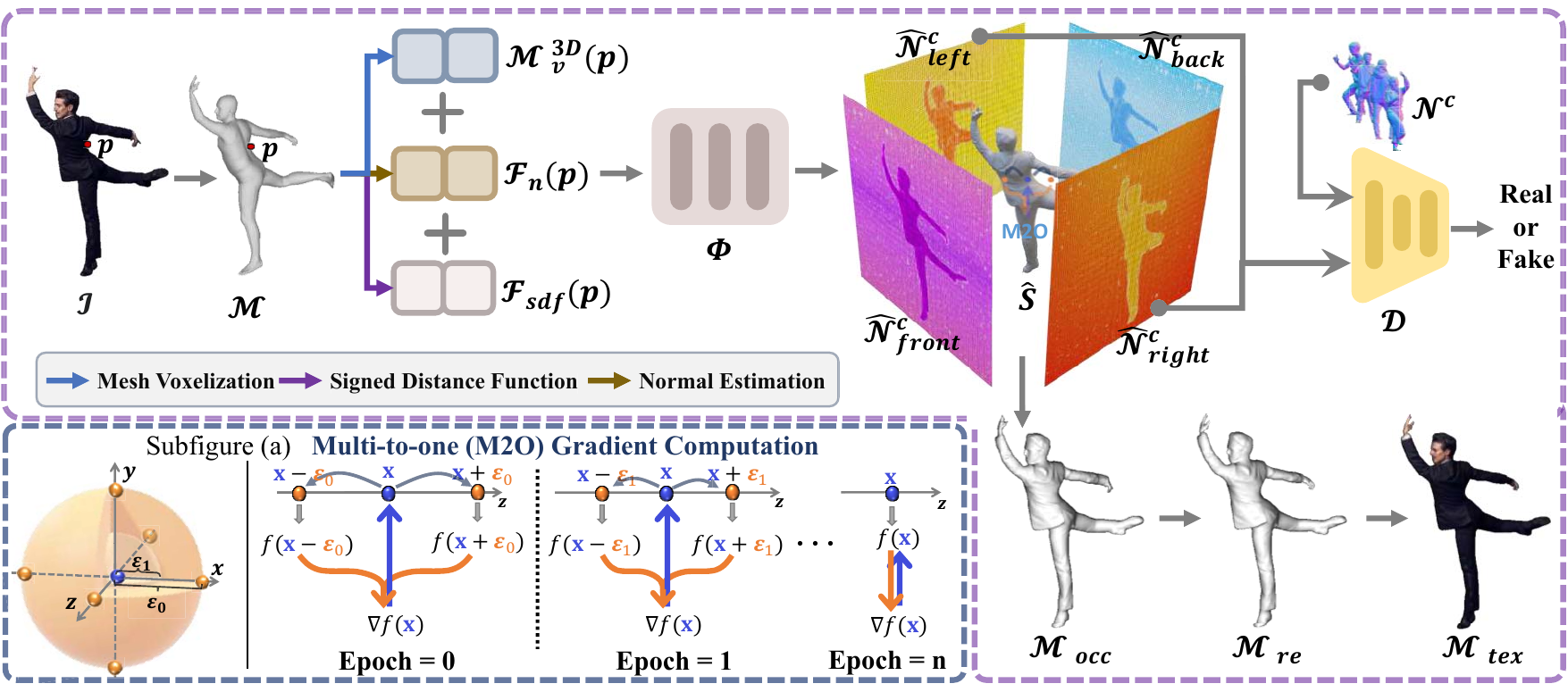}
    \caption{Overview of our \sexyname. 
    Conditioned on a single-view image $\mathcal{I}$ and the corresponding SMPL-X $\mathcal{M}$, we sample 3D points $\mathbf{P}$, obtaining their features regarding the geometry of the 3D clothed human. Then, we obtain the SDF $\hat{\mathcal{S}}$ and the corresponding four-view normal maps $\hat{\mathcal{N}}^c$ \wrt~the clothed human from $\mathcal{I}$. To improve the global side-view geometry, we introduce a~\ourdis $\mathcal{D}$ that distinguishes between $\hat{\mathcal{N}}_{side}^{c}= \{ \hat{N}_{left}^{c}, \hat{{N}}_{right}^{c}\}$ and ground truth $\mathcal{N}^c$. Moreover, to enhance local surface consistency, we employ a \ourng combined with a coarse-to-fine strategy. By integrating the gradients of nearby points around $p$, we smooth out local irregularities, leading to consistent surface reconstruction $\mathcal{M}_{occ}$. From subfigure (a), we sample two additional points along each axis (x, y, and z) of the canonical 3D coordinate system around a single point $\mathbf{p}_i$, with a step size $\epsilon$. The value of $\epsilon$ decreases as training proceeds.  Additionally, we can substitute the hands with SMPL-X models and map textures to mesh for enhanced visuals.}
    \label{fig:pipline}
\end{figure*}

We aim to reconstruct detailed 3D clothed avatars from single-view RGB images $\mathcal{I}$. 
Recent advances~\cite{yang2024hilo, zhang2024sifu, xiu2022icon} tend to use parametric model $\mathcal{M}$ such as SMPL-X~\cite{pavlakos2019smplx} to provide semantic regularization and use implicit function to infer the geometry of clothed avatars. We observe that these manners fail to fully regularize the side view, leading to inferior side view reconstruction results. 

To address this, our key idea is to regularize side-view geometry both on the global and local scale. As shown in Fig.~\ref{fig:pipline}, our proposed \sexyname ~contains two key components: 1) A~\ourdis that penalizes our implicit function if the generated side-view normal maps disagree with the real one; 2) A multi-to-one gradient computation (M2O) strategy that ensures a locally consistent geometry. M2O calculates the gradient of each sample point by integrating the gradients of its nearby points. 

Finally, we use the implicit function to obtain SDF $\mathcal{\hat{S}}$ by querying points $\mathbf{P}$ and leverage the Marching Cubes \cite{lorensen1998marching} to obtain the 3D mesh of the clothed avatar $\mathcal{\hat{M}}_{occ}$ from $\mathcal{\hat{S}}$. 

\subsection{\discriminator} \label{sec:VFD}
Recent advances usually fail to fully leverage side-view information of clothed humans, leading to inferior results (see Fig.~\ref{fig:vis_comp}). The input single image usually provides texture information about the front view, \ie~the $0^\circ$ angle that directly faces the camera, and the back view outline. We can leverage the outline of the front and the back view of the human body to constrain the silhouette of the reconstructed 3D human. However, the image provides limited texture and shape information regarding the side view. Recent methods~\cite{zhang2024sifu,yang2024hilo} use the SMPL-X model to provide the coarse side-view information of the naked body. However, this information lacks the necessary details on cloth and hair to regularize the human, resulting in global unreasonable geometries such as broken geometry. 
To address this, 
\ld{our key idea is to enhance the 3D reconstruction network using our designed discriminator that distinguishes between real and generated side-view normal maps.}\\
\textbf{Side-view Normal Maps Rendering}.
To obtain fake side-view normal maps to constrain the training of our \sexyname, we render reconstructed clothed human mesh $\mathcal{\hat{M}}_{occ}$ from two side views (\ie~$90^\circ$, $270^\circ$), resulting in $\hat{\mathcal{N}}_{side}^{c}= \{ \hat{N}_{left}^{c}, \hat{{N}}_{right}^{c}\}$. In our paper, we consider the camera viewpoint of the input image as the front view, \ie~with degree $0^{\circ}$. Additionally, for real normal maps $\mathcal{N}^{c}$,  we follow ICON~\cite{xiu2022icon} to obtain the front and back normal maps using the given single-view image and a pretrained normal prediction model. Note that we use intermediate data during the training process to obtain the real normal maps, so no additional cost is introduced. Moreover, $\hat{\mathcal{N}}_{side}^{c}$ and $\mathcal{N}^{c}$ do not necessarily come from the same individual, allowing us to expand more possible combinations to benefit the training process. We then introduce a discriminator to distinguish the real and fake normal maps.\\
\textbf{Patch-based Discriminator.} 
 It is essential to model high-frequency information to constrain our \sexyname~to generate 3D clothed humans with realistic geometry. The reason is that high-frequency normal maps not only contain reasonable geometry shapes but also include details like cloth wrinkles. To achieve this, we employ a patch-based discriminator~\cite{pix2pix2017} $\mathcal{D}$ that penalizes normal maps at the patch scale. Specifically, we evenly split our normal maps $\mathcal{N}^{c}$ and  $\hat{\mathcal{N}}_{side}^{c}$ into $3 \times 3$ patches and classify each patch in an image as either fake or real. We then average all responses for the final output of $\mathcal{D}$.\\
 \textbf{Adversarial Loss.}
Following the adversarial learning paradigm of \ld{GAN~\cite{isola2017image}}, we compute the adversarial loss via:
\begin{equation}
\begin{aligned}
\mathcal{L}_{\text{D}} = \mathcal{L}_{mse}(\mathcal{D}(\hat{\mathcal{N}}^c_{side}), 0) + \mathcal{L}_{mse}(\mathcal{D}(\mathcal{N}^c), 1).
\end{aligned}
\end{equation}
 
\subsection{Multi-to-one Gradient Computation} \label{sec:NG} 
Insufficient supervision from a single-view image results in a lack of local inconsistency, \eg~stitching and jagged edges in side views (see Fig.~\ref{fig:vis_comp}). To address this issue, we introduce a multi-to-one gradient computation that achieves a balance between smoothness and detailed reconstruction. 
\noindent \textbf{The Locality Issue of Eikonal constraint}.
Existing 3D reconstruction methods often achieve surface reconstruction by predicting the SDF of sampling points. Gropp \etal~\cite{gropp2020implicit} employ the Eikonal Constraint to ensure $\phi(\mathbf{P})$ from a point with coordinate $\mathbf{p}_i=(x_i, y_i, z_i)$ is a valid SDF using equation
$\mathcal{L}_{\text{eik}}(\nabla \phi(\mathbf{P})) = \frac{1}{N} \sum_{i=1}^{N} \left( \|\nabla \phi(\mathbf{p}_i)\|_2 - 1 \right)^2$,
where ${N}$ is the number of sampled points. Moreover, if one wants to further encourage the smoothness of the reconstructed surface, a solution is to constrain the second-order Eikonal gradient, namely the
curvature loss: 
$\mathcal{L}_{\text {curv }}(\nabla^2 \phi(\mathbf{P}))=\frac{1}{N} \sum_{i=1}^N\left|\nabla^2 \phi(\mathbf{p}_i)\right|$.
While the first-order and second-order Eikonal constraints are effective, the derivative $\nabla \phi(\mathbf{p}_i)$ and $\nabla^2 \phi(\mathbf{p}_i)$ are local, \textit{i.e.}, the  SDF of nearby points will be significantly different. 
For reconstruction, when sampling multiple nearby points, we expect these points to produce coherent SDF without abrupt transitions, thereby ensuring a locally smooth surface. \\
\textbf{Multi-to-one Gradient Computation.}
To overcome the locality issue of Eikonal constraint, we use Multi-to-one (\textbf{M2O}) gradients computation~\cite{li2023neuralangelo} that modifies $\nabla \phi(\mathbf{p}_i)$ and $\nabla^2 \phi(\mathbf{p}_i)$ to $\nabla \phi_{xyz}(\mathbf{p}_i)$ and $\nabla^2 \phi_{xyz}(\mathbf{p}_i)$. Intuitively, to ensure the local \ld{smoothness} of the reconstructed surface, we aim to minimize significant fluctuations in the gradients of nearby points. 

To achieve this, unlike existing methods that only backpropagate the gradient of a single point $\mathbf{p}_i$, we sample two additional points along each axis of the canonical coordinates around $\mathbf{p}_i$ in a step size $\epsilon$. 
The gradients of the additional points are integrated into the sampling points for backpropagation.  From Fig.~\ref{fig:pipline} (a), the z-component is
\begin{equation}
\begin{aligned}
\nabla_z\phi(\mathbf{p}_i)=\frac{\phi\left(\mathbf{p}_i+{\boldsymbol{\epsilon}}_z\right)-\phi\left(\mathbf{p}_i-\boldsymbol{\epsilon}_z\right)}{2 \epsilon}
\end{aligned}
\label{eq:eik}
\end{equation}
\begin{equation}
\begin{aligned}
\nabla_z^2 \phi(\mathbf{p}_i)=\frac{\phi\left(\mathbf{p}_i+\boldsymbol{\epsilon}_z\right)+\phi\left(\mathbf{p}_i-\boldsymbol{\epsilon}_z\right)-2\phi(\mathbf{p}_i)}{ \epsilon^2}
\end{aligned}
\label{eq:eik_2order}
\end{equation}
where $\boldsymbol{\epsilon}_z = [0, 0, \epsilon]$.  
Six additional samples is used for the surface normal computation for each sampling point. 

\begin{figure*}[th]
    \centering
    \includegraphics[width=1.0\linewidth]{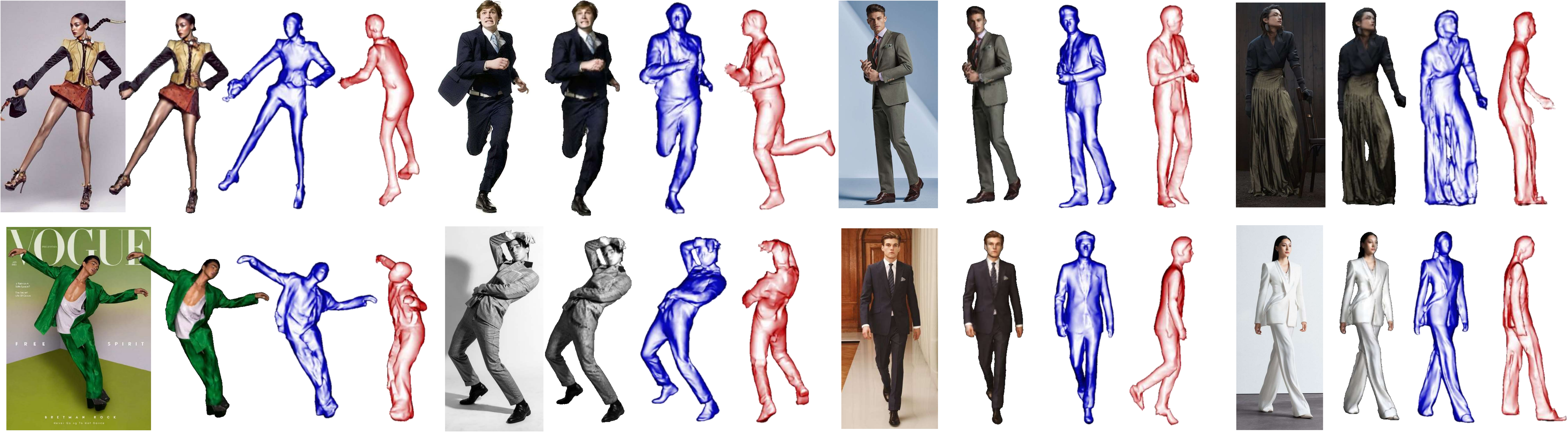}
    \caption{Visualization results of 3D clothed avatar reconstruction with our \sexyname~from in-the-wild images, which present various clothing and challenging poses. We show the front (\textcolor[RGB]{62,113,197}{blue}) and side (\textcolor[RGB]{195,46,46}{red}) views. Zoom in for more details.}
    \label{fig:inthewild}
\end{figure*}

\begin{figure}[t]
    \centering
    \includegraphics[width=0.45\textwidth]{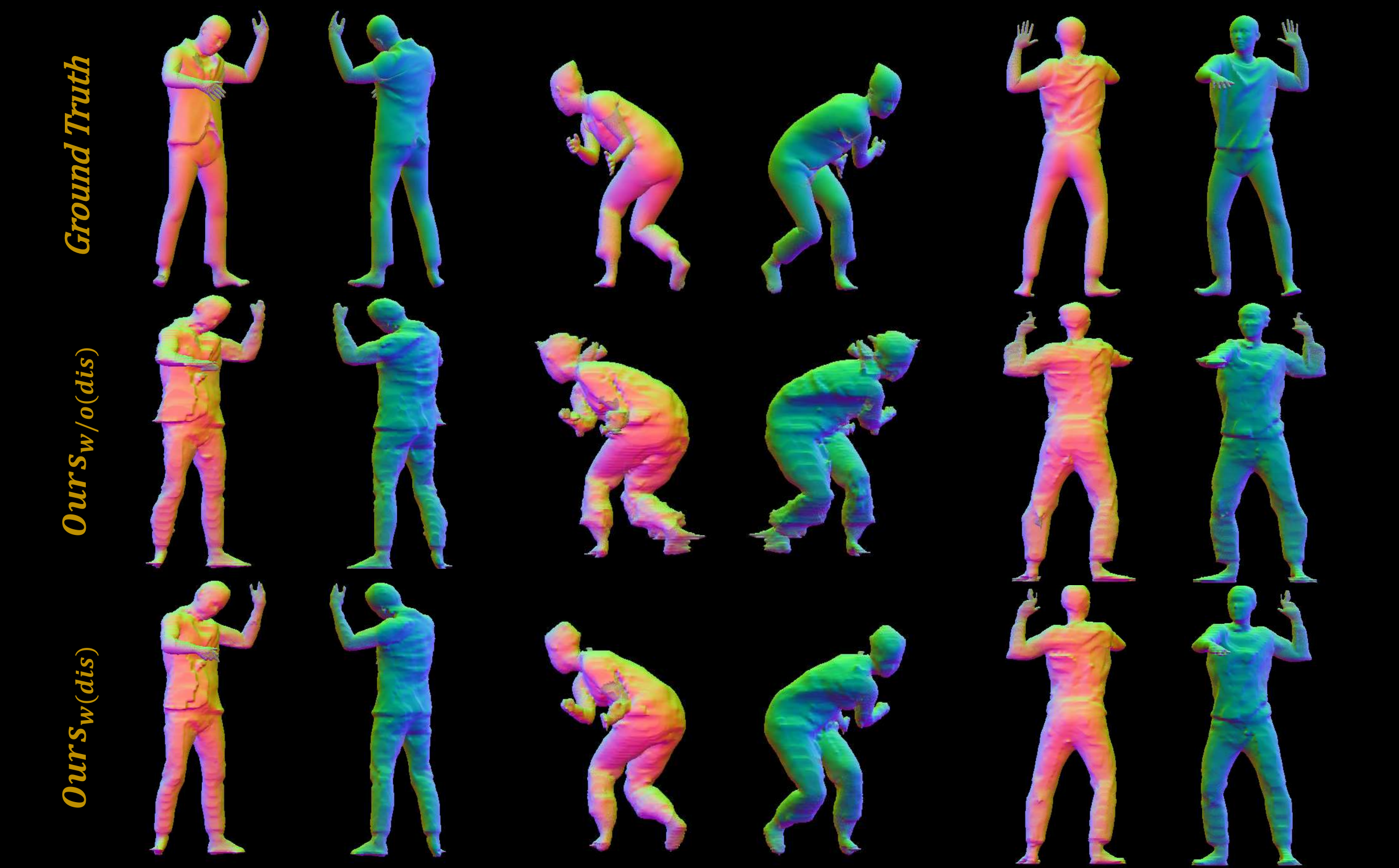}
    \caption{Reconstruction results on CAPE dataset with or without our~\ourdis. All the images are side normal maps relative to the input image. \ourdis can greatly enhance the geometry topology in the side view. The non-human parts are removed and the contours are more reasonable.} 
    \label{fig:ablation_dis}
\end{figure}

\begin{table}[t]
  \centering
 \caption{Comparison experiments and ablation studies on seen (\ie~training and test on the same dataset) and unseen (\ie~training on Thuman2.0 and test on CAPE) settings. The "-" denotes that the methods only release code for inference.}
     \begin{threeparttable}
 \small
\setlength{\tabcolsep}{3pt} 
\resizebox{1\linewidth}{!}
{
    \begin{tabular}{c|ccc|ccc|ccc}
    \toprule
Dataset & \multicolumn{3}{c|}{CAPE-FP} & \multicolumn{3}{c|}{CAPE-NFP} &\multicolumn{3}{c}{THuman2.0}\\
\hline
Methods & Chamfer ($\downarrow$) & P2S ($\downarrow$)  & Normals ($\downarrow$) & Chamfer & P2S & Normals & Chamfer & P2S   & Normals \\
\hline
          PIFu~\cite{saito2019pifu} & 2.1000  & 2.0930  & 0.0910  & 2.9730  & 2.9400  & 0.1110  & 2.6880  & 2.5730  & 0.0970  \\
          PIFuHD~\cite{saito2020pifuhd} &2.3020 &2.3350 &0.0900 &3.7040 &3.5170 &0.1230 &2.4613 &2.3605 &0.0924\\
          PaMIR~\cite{zheng2021pamir} & 1.2250  & 1.2060  & 0.0550  & 1.4130  & 1.3210  & 0.0630 & 1.4388 & 1.5613  & 0.1071 \\
\hline
          ICON~\cite{xiu2022icon}  & 0.7475  & 0.7488  & 0.0508  & 0.8656  & 0.8639  & 0.0545 & 1.1431  & 1.3020  & 0.0923 \\
          ECON~\cite{xiu2023econ}  & 0.9651  & 0.9175  & 0.0412  & 0.9983  & 0.9694  & 0.0415   & -  & -  & - \\
          D-IF~\cite{yang2023dif}  & 0.8038  & 0.7766  & 0.0546  & 0.9877  & 0.9491  & 0.0611 & 1.0305  & 1.0864  & 0.0830 \\
          HiLo~\cite{yang2024hilo}  & 0.7392  & 0.7299  & 0.0498  & 0.8620  & 0.8621  & 0.0541 & 0.9567  & 1.0380  & 0.0780 \\
        SIFU~\cite{zhang2024sifu}  & 0.7949  & 0.7624  & 0.0489  & 0.9170  & 0.8821  & 0.0530 & -  & -  & - \\
								
\hline

\cellcolor[rgb]{ .949,  .863,  .859}\sexyname~(Ours)  & \cellcolor[rgb]{ .949,  .863,  .859}0.7076 & \cellcolor[rgb]{ .949,  .863,  .859}0.6904 & \cellcolor[rgb]{ .949,  .863,  .859}0.0426 & \cellcolor[rgb]{ .949,  .863,  .859}0.8428 & \cellcolor[rgb]{ .949,  .863,  .859}0.8270 & \cellcolor[rgb]{ .949,  .863,  .859}0.0472 & \cellcolor[rgb]{ .949,  .863,  .859}0.9159 & \cellcolor[rgb]{ .949,  .863,  .859}0.9814 & \cellcolor[rgb]{ .949,  .863,  .859}0.0664\\

        $\mathrm{\sexyname}_{\mathrm{w/o}~ dis}$ & 0.7275  & 0.7127  & 0.0444  & 0.8518  & 0.8455  & 0.0486 & 0.9414  & 1.0183  & 0.0723 \\
        $\mathrm{\sexyname}_{\mathrm{w/o}~ m2o}$ & 0.7209 & 0.6949 & 0.0467 & 0.8793  & 0.8564   & 0.0516  & 1.0148 & 1.0587 & 0.0749 \\
    \bottomrule
    \end{tabular}%
}
    \end{threeparttable}
  \label{tab:comparison}%
\end{table}

\begin{figure}[t]
    \centering
    \includegraphics[width=0.45\textwidth]{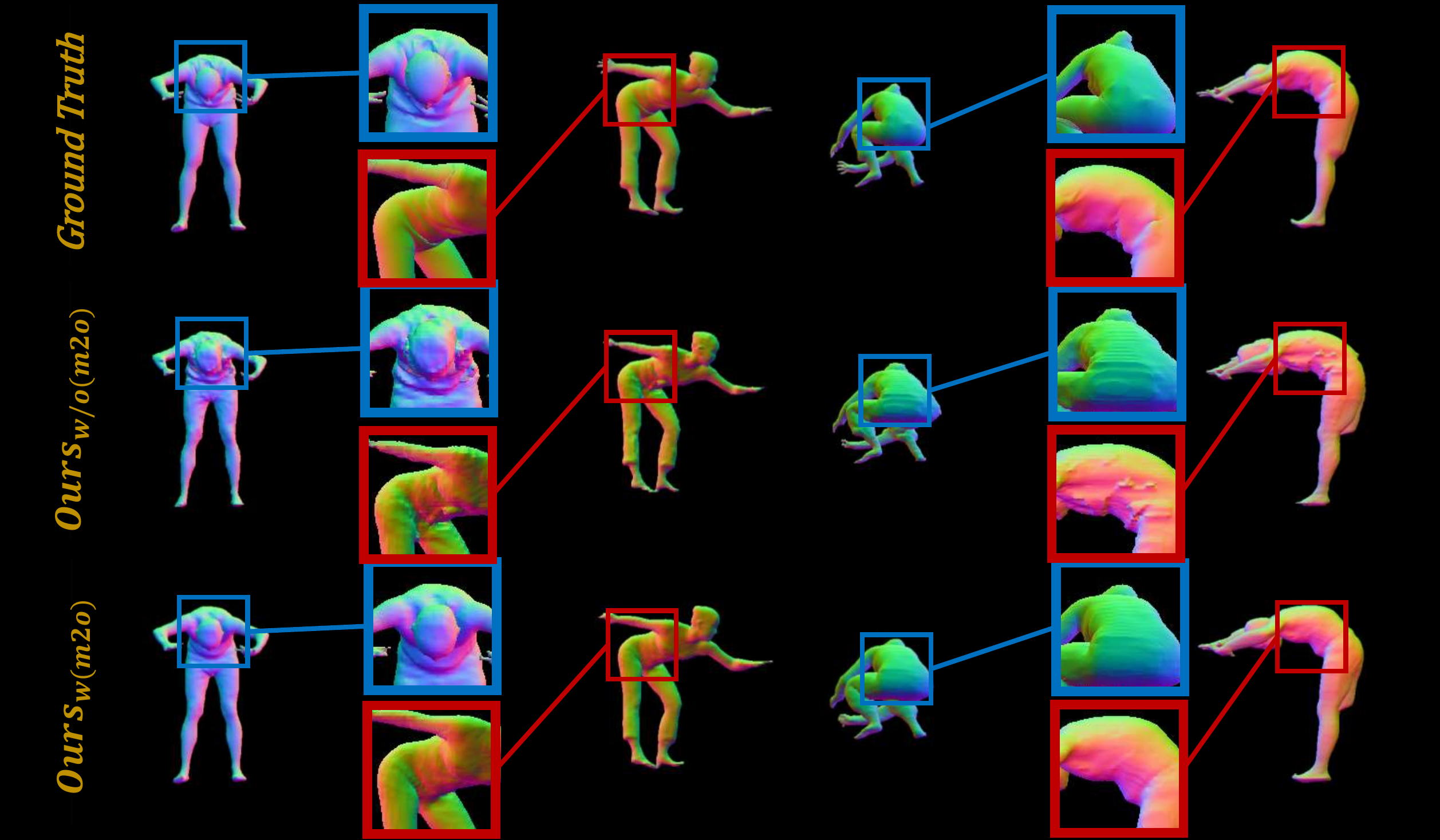}
    \caption{Reconstruction results on CAPE dataset with or without our \ourng. The results show that our \ourng can effectively eliminate the jagged surfaces and produce more consistent and more realistic geometry.}
    \label{fig:ablation_m2o}
\end{figure}

\noindent  \textbf{Coarse-to-fine Sampling.}
Reconstructing a high-quality 3D human requires balancing local surface smoothness and local details. To prevent the multi-to-one optimization from over-smoothing, we introduce coarse-to-fine sampling. 
Specifically, we exponentially decrease the scale of the step size $\epsilon$ as the training epoch $n$ proceeds. The coarse-to-fine procedure follows the equation:
    $\epsilon_n=\epsilon_0\cdot (1/2)^{n}$,
where $\epsilon_0=1/max(h,w)$ is the initial value, by setting this value we ensure the smooth range. h and w are the height and width of our rendered image.
From Fig.~\ref{fig:pipline}, during the early phase of the training such as epochs 0 and 1, $\epsilon_n$ is relatively large. The gradients of nearby points around the sampling point are integrated using Eqn.~(\ref{eq:eik}) and Eqn.~(\ref{eq:eik_2order}), which ensures the surface is smooth at a relatively large range. In the later phases of training, as $\epsilon_n$ becomes smaller, the gradient of a sampling point closely approximates its original gradient. This manner allows our \sexyname to capture high-frequency information about each point, thereby focusing on reconstructing finer geometric details.

\subsection{Optimization} \label{sec:optim} 
The total loss is defined as the weighted sum of the losses:
\begin{equation}
\begin{aligned}
\mathcal{L}=&\mathcal{L}_{a}+w_{\text {d}} \mathcal{L}_{D}+w_{\text {e}} \mathcal{L}_{\text {eik}}+w_{\text {c}} \mathcal{L}_{\text {curv}}
\end{aligned}
\end{equation}
where $\mathcal{L}_{a}=\mathcal{L}_{mse}(\mathcal{S}, \hat{\mathcal{S}})$ aligns the ground truth SDF $\mathcal{S}$ and the predicted one $\hat{\mathcal{S}}$. $w_{\text {d}}$,  $w_{\text {e}}$, and $w_{\text {c}}$ are hyper-parameters to balance different losses. All network parameters, including MLPs and Discriminator, are trained jointly.

\section{Experiment}
\noindent \textbf{Datasets:} 
We conduct experiments on two open-source datasets, \ie~THUman2.0~\cite{zheng2019deephuman} and CAPE~\cite{CAPEma2020learning}, which both contain various humans with different poses and diverse clothes. We train our model on the THuman2.0 dataset and test it on both datasets to further analyze our model’s generalization to complex body poses. Specifically, following ICON~\cite{xiu2022icon}, the CAPE dataset is divided into the "CAPE-FP" and "CAPE-NFP" sets, which have "fashion" and "non-fashion" poses, respectively. 
Moreover, to evaluate our \sexyname~on in-the-wild images, we follow ICON to collect 200 diverse images from Pinterest\footnote{https://www.pinterest.com/}. The images contain humans performing dramatic movements or wearing diverse clothes. 

\noindent \textbf{Metrics:} 
We evaluate \sexyname~and baselines in terms of three metrics. \textbf{Chamfer distance} and \textbf{P2S distance} measure reconstructed meshes with ground truth. \textbf{Normal error} measures normal images from both meshes. 

\noindent \textbf{Baselines:} 
We compare our proposed \sexyname~with mainstream state-of-the-art methods, including PIFu~\cite{saito2019pifu}, PIFuHD~\cite{saito2020pifuhd}, PaMIR~\cite{zheng2021pamir}, ICON~\cite{xiu2022icon}, ECON~\cite{xiu2023econ}, D-IF~\cite{yang2023dif}, HiLo~\cite{yang2024hilo} and SIFU~\cite{zhang2024sifu}.

For ablation studies, we construct two variants of our \sexyname: 
1) $\mathrm{\sexyname}_{{w/o}~dis}$ is constructed by removing our~\ourdis;
2) $\mathrm{\sexyname}_{{w/o}~m2o}$ is constructed by removing our Multi-to-one gradient computation.

\noindent \textbf{Implementation details:} 
We implement our approach using PyTorch
\cite{paszke2019pytorch} and train our networks with RMSprop

optimizer. For a fair comparison, we follow all common hyper-parameter settings the same as ICON~\cite{xiu2022icon}. 

\subsection{Comparison Experiments}
\textbf{Quantitative results.} We evaluate our \sexyname~in Tab.~\ref{tab:comparison} under two settings. 1) \emph{Setting1}: Following the setting of the previous methods, we train and test on the THuman2.0 datasets.  2) \emph{Setting2}: To further evaluate the generalization ability of our \sexyname~on unseen datasets, we train on THuman2.0 and test on CAPE dataset. To ensure a fair and reasonable comparison, all methods use the pretrained normal map prediction model to obtain normal maps instead of using the ground truth ones. Our \sexyname~achieves the best results across diverse scenarios due to the side-view consistent paradigm.  

\noindent \textbf{Visualization Results.}
We compare our \sexyname~with baselines on in-the-wild images in Fig.~\ref{fig:vis_comp}.
The results show that our \sexyname~is able to reconstruct 3D clothed avatars with more realistic details. Although ECON obtains detailed fingers by replacing the hand of the SMPL-X model, there exists misalignment on the connection wrist when the corresponding SMPL-X is inaccurate.
We put more visualization results of our \sexyname~on in-the-wild images in Fig.~\ref{fig:inthewild}.
The results show the effectiveness and generalization ability of our \sexyname~in recovering complex scenarios (loose clothing and challenging poses) and detailed geometry (such as hairs and wrinkles). 

\subsection{Ablation Studies} \label{sec:ablation}
\textbf{How does~\ourdis improve the body topology of the reconstructed avatar?}
As shown in Fig.~\ref{fig:ablation_dis} and Tab.~\ref{tab:comparison}, our \ourdis removes the non-human shape and boosts reconstruction performance. The reason is that our \ourdis emphasizes the rationality of the lateral normal maps in the training process, forcing the generation of geometric structures closer to the real situation. 

\noindent \textbf{How does Multi-to-one Gradient Computation (M2O) improve geometry consistency?}
We quantitatively demonstrate the necessity of M2O in Tab.~\ref{tab:comparison}. The results demonstrate that $\mathrm{\sexyname}_{{w/o}~m2o}$~ achieves inferior performance than \sexyname. Furthermore, Fig.~\ref{fig:ablation_m2o} demonstrates that M2O leads to more consistent reconstruction, resulting in more realistic surfaces. From the results, we observe that incorporating our M2O helps in capturing consistent geometry. 

\begin{table}[th]
  \centering
        \caption{Evaluation of contributions of each component in our model on the CAPE-NFP dataset.}
     \begin{threeparttable}
 \small
\setlength{\tabcolsep}{3pt} 
\resizebox{0.65\linewidth}{!}
{
    \begin{tabular}{c|ccc}
    \toprule
Dataset &\multicolumn{3}{c}{CAPE-NFP}\\
\hline
Methods & Chamfer ($\downarrow$) & P2S ($\downarrow$)  & Normals ($\downarrow$) \\
\hline
\multicolumn{4}{c}{\textit{A~-~Different Discriminator settings}}\\
\hline
    BCE Loss & 0.9568  & 0.9416  & 0.0689\\
        Vanilla GAN & 0.9940  & 0.9725  & 0.0701\\
\hline
\multicolumn{4}{c}{\textit{B~-~Different Normal Map Number}}\\
\hline
        4 Normal Maps   & 0.9580  & 0.9228  & 0.0576\\
\hline
\multicolumn{4}{c}{\textit{C~-~Different Sampling Step}}\\
\hline
        w/o C2F Sampling & 1.0210  & 0.9942  & 0.0561\\
\hline
          \cellcolor[rgb]{ .949,  .863,  .859}\sexyname~(Ours) & \cellcolor[rgb]{ .949,  .863,  .859}0.8428  & \cellcolor[rgb]{ .949,  .863,  .859}0.8270  & \cellcolor[rgb]{ .949,  .863,  .859}0.0472\\
    \bottomrule
    \end{tabular}%
}
    \end{threeparttable}
  \label{tab:mse}%
\end{table}

\subsection{More discussion}
\label{sec:discussion}
\noindent  \textbf{How does our Discriminator work?}
We analyze the backbone and loss function of our \ourdis. As shown in Tab.\ref{tab:mse}, since the vanilla GAN\cite{NIPS2014gan} discriminator fails to fully exploit the high-frequency information in normal maps, it achieves inferior performance. In contrast, our patch-based backbone is better suited for extracting high-frequency information. Furthermore, the MSE loss outperforms the BCE loss in producing superior results. Based on the analysis, we select the patch-based backbone and MSE loss for our \ourdis.

\noindent  \textbf{How does the number of normal maps affect \sexyname?}
We explore utilizing our discriminator to differentiate between the four normal map views (i.e., front, back, left, and right). However, as shown in Tab.~\ref{tab:mse}, incorporating all four views degrades performance. 
The primary reason is that the input image already provides sufficient front texture and back contour. Additional information from the normal maps may disrupt the training process, leading to suboptimal results. Based on this, we use the two side-view normal maps for the discriminator.

\noindent \textbf{How does the coarse-to-fine sampling affect the reconstructed human?}
The step size $\epsilon$ of \textbf{M2O} controls the range of points for integrating gradients. A larger $\epsilon$ ensures the surface is smooth at a larger scale, while a smaller $\epsilon$ affects a smaller region and encourages geometry to be detailed. We propose a coarse-to-fine sampling to achieve a balance between smoothness and details. From Tab.~\ref{tab:comparison}, \ld{our \textbf{M2O} using coarse-to-fine $\epsilon$ outperforms \textbf{M2O} using fixed $\epsilon$.}


\section{Conclusion}
We introduce \sexyname, a novel method for reconstructing high-quality 3D meshes of clothed humans with realistic and accurate surfaces, especially in side views. Our approach employs a patch-based normal discriminator to refine side-view geometry, enhancing the global human topology in side views. Moreover, we propose a multi-to-one gradient computation that integrates gradients of nearby points for a sampled point, ensuring local surface consistency. \sexyname outperforms existing methods in both quantitative and qualitative evaluations, demonstrating exceptional capability in handling diverse poses and clothing, making it ideal for applications like animation.

\section*{Acknowledgment}
This work was partially supported by the Joint Funds of the National Natural Science Foundation of China (Grant No.U24A20327), the Funding by Science and Technology Projects in Guangzhou (2023A04J1686), and TCL Science and Technology Innovation Fund, China.

\bibliographystyle{IEEEbib}
\bibliography{icme2025references}

\begin{thebibliography}{10}

\bibitem{yoon2019effect}
Boram Yoon, Hyung-il Kim, Gun~A Lee, Mark Billinghurst, and Woontack Woo,
\newblock ``The effect of avatar appearance on social presence in an augmented
  reality remote collaboration,''
\newblock in {\em IEEE Conference on Virtual Reality and 3D User Interfaces
  (VR)}. IEEE, 2019, pp. 547--556.

\bibitem{sun2022research}
Lin Sun,
\newblock ``Research on the application of 3d animation special effects in
  animated films: Taking the film avatar as an example.,''
\newblock {\em Scientific Programming}, 2022.

\bibitem{schonberger2016pixelwise}
Johannes~L Sch{\"o}nberger, Enliang Zheng, Jan-Michael Frahm, and Marc
  Pollefeys,
\newblock ``Pixelwise view selection for unstructured multi-view stereo,''
\newblock in {\em ECCV}, 2016.

\bibitem{guo2019relightables}
Kaiwen Guo, Peter Lincoln, Philip Davidson, Jay Busch, Xueming Yu, Matt Whalen,
  Geoff Harvey, Sergio Orts-Escolano, Rohit Pandey, Jason Dourgarian, et~al.,
\newblock ``The relightables: Volumetric performance capture of humans with
  realistic relighting,''
\newblock {\em ACM Transactions on Graphics}, vol. 38, no. 6, pp. 1--19, 2019.

\bibitem{zheng2014patchmatch}
Enliang Zheng, Enrique Dunn, Vladimir Jojic, and Jan-Michael Frahm,
\newblock ``Patchmatch based joint view selection and depthmap estimation,''
\newblock in {\em CVPR}, 2014, pp. 1510--1517.

\bibitem{alldieck2019tex2shape}
Thiemo Alldieck, Gerard Pons-Moll, Christian Theobalt, and Marcus Magnor,
\newblock ``Tex2shape: Detailed full human body geometry from a single image,''
\newblock in {\em CVPR}, 2019, pp. 2293--2303.

\bibitem{saito2019pifu}
Shunsuke Saito, Zeng Huang, Ryota Natsume, Shigeo Morishima, Angjoo Kanazawa,
  and Hao Li,
\newblock ``Pifu: Pixel-aligned implicit function for high-resolution clothed
  human digitization,''
\newblock in {\em ICCV}, 2019, pp. 2304--2314.

\bibitem{saito2020pifuhd}
Shunsuke Saito, Tomas Simon, Jason Saragih, and Hanbyul Joo,
\newblock ``Pifuhd: Multi-level pixel-aligned implicit function for
  high-resolution 3d human digitization,''
\newblock in {\em CVPR}, 2020, pp. 84--93.

\bibitem{zheng2021pamir}
Zerong Zheng, Tao Yu, Yebin Liu, and Qionghai Dai,
\newblock ``Pamir: Parametric model-conditioned implicit representation for
  image-based human reconstruction,''
\newblock {\em IEEE TPAMI}, vol. 44, no. 6, pp. 3170--3184, 2021.

\bibitem{xiu2022icon}
Yuliang Xiu, Jinlong Yang, Dimitrios Tzionas, and Michael~J. Black,
\newblock ``{ICON}: {I}mplicit {C}lothed humans {O}btained from {N}ormals,''
\newblock in {\em CVPR}, 2022, pp. 13296--13306.

\bibitem{xiu2023econ}
Yuliang Xiu, Jinlong Yang, Xu~Cao, Dimitrios Tzionas, and Michael~J. Black,
\newblock ``{ECON: Explicit Clothed humans Optimized via Normal integration},''
\newblock in {\em CVPR}, June 2023.

\bibitem{yang2023dif}
Xueting Yang, Yihao Luo, Yuliang Xiu, Wei Wang, Hao Xu, and Zhaoxin Fan,
\newblock ``D-if: Uncertainty-aware human digitization via implicit
  distribution field,''
\newblock in {\em ICCV}, 2023, pp. 9122--9132.

\bibitem{xiang2020monoclothcap}
Donglai Xiang, Fabian Prada, Chenglei Wu, and Jessica Hodgins,
\newblock ``Monoclothcap: Towards temporally coherent clothing capture from
  monocular rgb video,''
\newblock in {\em International Conference on 3D Vision (3DV)}. IEEE, 2020, pp.
  322--332.

\bibitem{loper2015smpl}
Matthew Loper, Naureen Mahmood, Javier Romero, Gerard Pons-Moll, and Michael~J
  Black,
\newblock ``Smpl: A skinned multi-person linear model,''
\newblock {\em ACM Transactions on Graphics}, vol. 34, 2015.

\bibitem{pavlakos2019smplx}
Georgios Pavlakos, Vasileios Choutas, Nima Ghorbani, Timo Bolkart, Ahmed~AA
  Osman, Dimitrios Tzionas, and Michael~J Black,
\newblock ``Expressive body capture: 3d hands, face, and body from a single
  image,''
\newblock in {\em CVPR}, 2019, pp. 10975--10985.

\bibitem{yang2024hilo}
Yifan Yang, Dong Liu, Shuhai Zhang, Zeshuai Deng, Zixiong Huang, and Mingkui
  Tan,
\newblock ``Hilo: Detailed and robust 3d clothed human reconstruction with
  high-and low-frequency information of parametric models,''
\newblock in {\em CVPR}, 2024, pp. 10671--10681.

\bibitem{zhang2024sifu}
Zechuan Zhang, Zongxin Yang, and Yi~Yang,
\newblock ``Sifu: Side-view conditioned implicit function for real-world usable
  clothed human reconstruction,''
\newblock in {\em CVPR}, 2024, pp. 9936--9947.

\bibitem{park2019deepsdf}
Jeong~Joon Park, Peter Florence, Julian Straub, Richard Newcombe, and Steven
  Lovegrove,
\newblock ``Deepsdf: Learning continuous signed distance functions for shape
  representation,''
\newblock in {\em CVPR}, 2019, pp. 165--174.

\bibitem{lorensen1998marching}
William~E Lorensen and Harvey~E Cline,
\newblock ``Marching cubes: A high resolution 3d surface construction
  algorithm,''
\newblock in {\em Seminal graphics: pioneering efforts that shaped the field},
  1998, pp. 347--353.

\bibitem{pix2pix2017}
Phillip Isola, Jun-Yan Zhu, Tinghui Zhou, and Alexei~A Efros,
\newblock ``Image-to-image translation with conditional adversarial networks,''
\newblock in {\em CVPR}, 2017.

\bibitem{isola2017image}
Phillip Isola, Jun-Yan Zhu, Tinghui Zhou, and Alexei~A Efros,
\newblock ``Image-to-image translation with conditional adversarial networks,''
\newblock in {\em CVPR}, 2017, pp. 1125--1134.

\bibitem{gropp2020implicit}
Amos Gropp, Lior Yariv, Niv Haim, Matan Atzmon, and Yaron Lipman,
\newblock ``Implicit geometric regularization for learning shapes,''
\newblock {\em ICML}, 2020.

\bibitem{li2023neuralangelo}
Zhaoshuo Li, Thomas M{\"u}ller, Alex Evans, Russell~H Taylor, Mathias Unberath,
  Ming-Yu Liu, and Chen-Hsuan Lin,
\newblock ``Neuralangelo: High-fidelity neural surface reconstruction,''
\newblock in {\em CVPR}, 2023, pp. 8456--8465.

\bibitem{zheng2019deephuman}
Zerong Zheng, Tao Yu, Yixuan Wei, Qionghai Dai, and Yebin Liu,
\newblock ``Deephuman: 3d human reconstruction from a single image,''
\newblock in {\em ICCV}, 2019, pp. 7739--7749.

\bibitem{CAPEma2020learning}
Qianli Ma, Jinlong Yang, Anurag Ranjan, Sergi Pujades, Gerard Pons-Moll, Siyu
  Tang, and Michael~J Black,
\newblock ``Learning to dress 3d people in generative clothing,''
\newblock in {\em CVPR}, 2020, pp. 6469--6478.

\bibitem{paszke2019pytorch}
Adam Paszke, Sam Gross, Francisco Massa, Adam Lerer, James Bradbury, Gregory
  Chanan, Trevor Killeen, Zeming Lin, Natalia Gimelshein, Luca Antiga, et~al.,
\newblock ``Pytorch: An imperative style, high-performance deep learning
  library,''
\newblock {\em NeurIPS}, vol. 32, 2019.

\bibitem{NIPS2014gan}
Ian Goodfellow, Jean Pouget-Abadie, Mehdi Mirza, Bing Xu, David Warde-Farley,
  Sherjil Ozair, Aaron Courville, and Yoshua Bengio,
\newblock ``Generative adversarial nets,''
\newblock in {\em NeurIPS}, 2014, vol.~27.

\end{thebibliography}

\end{document}